\documentclass[sigconf]{acmart}

\usepackage{amssymb,amsmath,amsfonts}
\usepackage{algorithm}
\usepackage[position=bottom,caption=false,font=footnotesize,labelfont=rm,textfont=rm]{subfig}
\usepackage{textcomp}
\usepackage{stfloats}
\usepackage{graphicx}
\usepackage{float}
\usepackage{multirow}
\usepackage{makecell}
\usepackage{color} 
\usepackage{hyperref}
\usepackage{bbding}
\usepackage{balance}

\usepackage{diagbox}
\usepackage[skins]{tcolorbox} 
\tcbuselibrary{breakable} 
\usepackage{paralist} 

\AtBeginDocument{%
  }

\copyrightyear{2025}
\acmYear{2025}
\setcopyright{acmlicensed}
\acmConference[MM '25] {Proceedings of the 33rd ACM International Conference on Multimedia}{October 27--31, 2025}{Dublin, Ireland.}
\acmBooktitle{Proceedings of the 33rd ACM International Conference on Multimedia (MM '25), October 27--31, 2025, Dublin, Ireland}
\acmISBN{979-8-4007-2035-2/2025/10}
\acmDOI{10.1145/3746027.3762029}

\begin{document}

\title{E3RG: Building Explicit Emotion-driven Empathetic Response Generation System with Multimodal Large Language Model}




\author{Ronghao Lin}
\orcid{0000-0003-4530-4529}
\affiliation{%
  \institution{Sun Yat-sen University}
  \city{Guangzhou}
  \state{Guangdong}
  \country{China}}
\affiliation{%
  \institution{Nanyang Technological University}
  \country{Singapore}}
\email{linrh7@mail2.sysu.edu.cn}

\author{Shuai Shen}
\orcid{0009-0005-9051-1107}
\affiliation{%
  \institution{Nanyang Technological University}
  \country{Singapore}}
\email{shuai.shen@ntu.edu.sg}

\author{Weipeng Hu}
\orcid{0000-0003-2886-7346}
\affiliation{%
  \institution{Nanyang Technological University}
  \country{Singapore}}
\email{weipeng.hu@ntu.edu.sg}

\author{Qiaolin He}
\orcid{0009-0001-2204-8668}
\affiliation{%
  \institution{Sun Yat-sen University}
  \city{Guangzhou}
  \state{Guangdong}
  \country{China}}
\email{heqlin5@mail2.sysu.edu.cn}

\author{Aolin Xiong}
\orcid{0009-0005-2301-7897}
\affiliation{%
  \institution{Sun Yat-sen University}
  \city{Guangzhou}
  \state{Guangdong}
  \country{China}}
\email{xiongaolin@mail2.sysu.edu.cn}

\author{Li Huang}
\orcid{0000-0001-8396-5243}
\affiliation{%
  \institution{Desay SV Automotive Co., Ltd}
  \city{Huizhou}
  \state{Guangdong}
  \country{China}}
\email{Li.Huang@desaysv.com}

\author{Haifeng Hu}
\orcid{0000-0002-4884-323X}
\affiliation{%
  \institution{Sun Yat-sen University}
  \city{Guangzhou}
  \state{Guangdong}
  \country{China}}
\affiliation{%
  \institution{Pazhou Laboratory}
  \city{Guangzhou}
  \state{Guangdong}
  \country{China}}
\email{huhaif@mail.sysu.edu.cn}

\author{Yap-peng Tan}
\orcid{0000-0002-0645-9109}
\affiliation{%
  \institution{Nanyang Technological University}
  \country{Singapore}}
\email{eyptan@ntu.edu.sg}

\renewcommand{\shortauthors}{Ronghao Lin et  al.}

\begin{abstract}
Multimodal Empathetic Response Generation (MERG) is crucial for building emotionally intelligent human-computer interactions. Although large language models (LLMs) have improved text-based ERG, challenges remain in handling multimodal emotional content and maintaining identity consistency. Thus, we propose E3RG, an Explicit Emotion-driven Empathetic Response
Generation System based on multimodal LLMs which decomposes MERG task into three parts: multimodal empathy understanding, empathy memory retrieval, and multimodal response generation. By integrating advanced expressive speech and video generative models, E3RG delivers natural, emotionally rich, and identity-consistent responses without extra training. Experiments validate the superiority of our system on both zero-shot and few-shot settings, securing Top-1 position in the Avatar-based Multimodal Empathy Challenge on ACM MM’25. Our code is available at \url{https://github.com/RH-Lin/E3RG}.

\end{abstract}




\maketitle

\section{Introduction}
\label{sec:introduction}
Emotional intelligence has become a vital aspect on the journey to Artificial General Intelligence (AGI), as it enhances human-like cognitive abilities by allowing machines to perceive, interpret, and respond to human emotions \cite{huang2024apathetic, sabour2024emobench,chen2024emotionqueen}. In this field, Empathetic Response Generation (ERG) has emerged as a challenging task in natural language processing, aiming to develop conversational systems capable of engaging in emotionally aware and contextually appropriate dialogue \cite{ma2020survey}.

To achieve a more comprehensive understanding of human intent and affect, the ERG task has evolved from text-only settings \cite{rashkin2019empatheticdialogues} to multimodal dialogue scenarios that better simulate real-world interactions \cite{zhang2025avamerg}. Multimodal Empathetic Response Generation (MERG) is designed with two core objectives: (1) to accurately understand emotions conveyed through both verbal and nonverbal cues, and (2) to generate nuanced, expressive video responses that align with the emotional and contextual dynamics of the dialogue.

Since large language models (LLMs) have demonstrated strong zero-shot transferability and robust semantic understanding, recent advancements have leveraged their capabilities to significantly improve performance in empathetic response generation (ERG) \cite{liu2021esconv}. By harnessing LLMs’ ability to comprehend nuanced context \cite{lian2025affectgpt}, these works have enhanced the coherence, relevance, and emotional alignment of generated responses in ERG tasks.

Therefore, the inherent empathetic capabilities of LLMs are intuitively extended to Multimodal Large Language Models (MLLMs), which have shown promising effectiveness in the MERG task \cite{qian2023harnessing}. However, as illustrated in Table \ref{table_method}, prior methods often rely on heavy post-training and elaborate fine-tuning strategies to enhance emotion understanding and empathetic video generation \cite{fu2023ecore, zhang2024stickerconv, fei2024empathyear}. These approaches are not only computationally expensive but also risk generalization limitation across diverse scenarios.

Moreover, the incorporation of multimodal video context introduces additional challenges in maintaining multimodal alignment and output consistency. To generate natural while semantically coherent talking-head responses, MERG systems must effectively synchronize emotional cues across diverse modalities \cite{yan2024perceptiveagent, zhang2025avamerg}. In addition, recent work such as PERGM \cite{huang2024pergm} emphasizes the role of specific personal profiles in delivering contextually appropriate empathy, highlighting the importance of identity consistency among the generated outputs and the talker's historical dialogue style.

In addition, although some prior methods consider the emotions of both speaker and listener during dialogues \cite{fu2023ecore,huang2024pergm,fei2024empathyear}, the challenge of modeling emotional multimodal context remains insufficiently addressed, particularly in the domains of expressive speech synthesis and avatar-based video generation \cite{li2023styletts, zhang2023sadtalker}. Existing systems often struggle to accurately capture and reproduce steady emotional dynamics across modalities, limiting the empathetic quality of generated responses.

To address the above issues, we propose an Explicit Emotion-driven Empathetic Response Generation system, named E3RG, built upon Multimodal Large Language Models (MLLMs). Specifically, the proposed E3RG is designed to achieve three key capabilities: emotion-awareness, personality-awareness, and knowledge-accessibility, which collectively enhance the expressiveness, coherence, and naturalness of the generated multimodal responses. By integrating the state-of-the-art generative models such as OpenVoice \cite{qin2023openvoice} for expressive speech synthesis and DICE-Talk \cite{tan2025dicetalk} for emotional talking-head generation, our system secures the Top-1 position in the Grand Challenge of Avatar-based Multimodal Empathetic Conversation at ACM MM’25. The main contributions of our approach are summarized as follows:

\begin{itemize}
    \item By decomposing the MERG task into three sub-tasks: multimodal empathy understanding, empathy memory retrieval, and multimodal empathy generation, the proposed E3RG system constructs a unified understanding and generation framework, which is designed with modular flexibility, allowing each component to be independently replaced or upgraded, thus ensuring adaptability to evolving model advancements and specific application needs.
    
    \item Deployed in a training-free manner, the MLLMs and expressive generative models integrated into our system are explicitly driven by emotion to achieve notable improvements in both zero-shot and few-shot scenarios. 

    \item Extensive experiments show the superiority of our system, reaching state-of-the-art performance by $76.3\%$ on HIT rate, $0.990$ on Dist-1 and average score $4.03$ on human evaluation.
\end{itemize}

\begin{table}[htbp]
\vspace{-0.1cm}
\caption{Comparison on diverse aspects of ERG system.}
\label{table_method}
\centering
\scalebox{0.8}{
\setlength\tabcolsep{5pt}
\begin{tabular}{c|cccc}

\hline
\begin{tabular}[c]{@{}c@{}}{System / Aspect}\end{tabular} &
\begin{tabular}[c]{@{}c@{}}Training-\\free\end{tabular} &
\begin{tabular}[c]{@{}c@{}}Emotion\\Guidance\end{tabular} &
\begin{tabular}[c]{@{}c@{}}Identity\\Consistency\end{tabular} &
\begin{tabular}[c]{@{}c@{}}Multimodal\\Video Context\end{tabular} \\
\hline


  PEGS \cite{zhang2024stickerconv} & \XSolidBrush & \XSolidBrush & \XSolidBrush & \XSolidBrush \\
  E-CORE \cite{fu2023ecore} & \XSolidBrush & \Checkmark & \XSolidBrush & \XSolidBrush \\
  PerceptiveAgent \cite{yan2024perceptiveagent} & \XSolidBrush & \XSolidBrush & \XSolidBrush & \Checkmark \\
  PERGM \cite{huang2024pergm} & \XSolidBrush & \Checkmark & \Checkmark & \XSolidBrush \\
  Empatheia \cite{zhang2025avamerg} & \XSolidBrush & \XSolidBrush & \Checkmark & \Checkmark \\
  EmpathyEar \cite{fei2024empathyear} & \XSolidBrush & \Checkmark & \Checkmark & \Checkmark \\
\hline
   \textbf{E3RG} & \Checkmark & \Checkmark & \Checkmark & \Checkmark \\
\hline
\end{tabular}
}
\vspace{-0.1cm}
\end{table}

\section{Related Work}
\label{sec:relatedwork}
\subsection{Multimodal Empathetic Response}
In the field of human-computer interaction, Empathetic Response Generation (ERG) has emerged as a cornerstone of affective computing \cite{ma2020survey,liu2021esconv,qian2023harnessing}, aiming to construct a conversational system with the capacity to recognize, interpret, and appropriately respond to humans with appropriate emotion, known as empathy \cite{raamkumar2023empathetic}. By fostering emotional consistency and providing harmonious support, empathetic dialogue systems can greatly enhance human users' satisfaction, trust, and engagement in wide applications ranging from customer service \cite{lehnert2024customer}, social media communication \cite{zhou2020social}, education \cite{sorin2024large} to mental health support \cite{qiu2024smile}. 
The ability to convey genuine understanding not only improves the naturalness and realism of human-computer interactions but also opens avenues for more effective intervention in domains where emotional sensitivity is primary \cite{raamkumar2023empathetic}. Consequently, advancing the quality and diversity of ERG remains an essential goal for the real‑world deployment of human-centric artificial intelligence \cite{huang2024apathetic}.

Initial efforts in ERG focused exclusively on linguistic utterance \cite{rashkin2019empatheticdialogues,liu2021esconv}, which greatly limits their real‑world applicability since natural human dialogue typically encompasses multiple modalities. Recent researches \cite{zhang2024stickerconv,yan2024perceptiveagent,zhang2025avamerg,fei2024empathyear} have devoted to integrating audio cues (e.g., pitch, frequency, tone) and visual signals (e.g., facial expression, gaze, body movement) alongside textual information to understand of users' emotion and produce more precise multimodal responses, named as Multimodal Empathetic Response Generation (MERG) \cite{fei2024empathyear,zhang2025avamerg}. 
Such multimodal understanding of humans' behavior and intent is essential in raising new perspectives of emotional intelligence in cognitive science \cite{zhao2024matter,sabour2024emobench}.

Moreover, beyond perceiving the multimodal input, MERG takes another step on the generated responses from text solely to multimodal outputs, including linguistic utterance, speech, and facial video \cite{fei2024empathyear,zhang2025avamerg}. With the emergence of auto-regressive model \cite{brown2020gpt3} and diffusion-based generative model \cite{ho2020ddpm}, recent multimodal systems have begun to couple diverse modality-specific generators to produce multiple unimodal responses by cross-modal interaction individually. However, maintaining multimodal consistency and semantic relevance are intuitively difficult across diverse modalities. Besides, emotion variations may accumulate during separate generation stages, ultimately undermining the system's overall empathetic quality.  Therefore, the key challenge for MERG lies in building a unified framework capable of generating contextual text, natural speech, and expressive talking-head videos in a coherent and emotionally synchronized manner.

\subsection{Multimodal Large Language Model}
Recent advances in Large Language Models (LLMs) have demonstrated remarkable abilities in language understanding, reasoning, and instruction following \cite{brown2020gpt3}. Building on these capabilities, research has increasingly moved toward Multimodal Large Language Models (MLLMs), which extend LLMs to process and integrate multimodal inputs, such as text, images, audio, and video \cite{zhang2024mmllm}. 
Recent studies explore joint multimodal learning framework with LLMs to understand modality-shared information and capture cross-modal dynamics \cite{lu2024unifiedio2,jin2025qwen25omni,liu2025ola}. 
Despite these advancements, most MLLMs focused on perception tasks, neglecting emotional understanding or generation with expressive multimodal outputs \cite{sabour2024emobench,huang2024apathetic,lin2024end,lian2025affectgpt}. 

In the context of empathetic response generation (ERG), most existing MLLMs fall short in generating coordinated multimodal outputs such as expressive speech and facial animations. Multimodal systems like NExT-GPT \cite{wu2024nextgpt}, VILA-U \cite{wu2025vilau}, and Janus \cite{wu2025janus} have made progress in general-purpose multimodal understanding and generation, but lack emotion-specific tuning and human-centric components like speech or talking-head generators, limiting their use in MERG scenarios \cite{chen2024emotionqueen}. 

Therefore, we aim to develop an effective MLLM-based system to flexibly handle both understanding and generation task in MERG, incorporating cross-modal alignment techniques and emotion-driven pipeline to produce coherent and expressive responses across text, speech, and facial video modalities.

\subsection{Expressive Text-to-Speech Generation}
Current Text-to-Speech (TTS) models have made rapid advancements following the advent of diverse types of generative models \cite{chen2025valle,wang2025maskgct,jiang2025sparse}. 
However, beyond naturalness and zero-shot robustness, generating expressive speech that considers prosody, emotion, and speaking styles still remains a challenge \cite{liu2023emotionally}. To address this, expressive TTS approaches have introduced emotion as a conditioning signal, such as Global Style Tokens \cite{wang2018style} which enables unsupervised learning to model the expressiveness and speaking style in global embeddings. More recent works, such as StyleTTS \cite{li2023styletts}, CosyVoice \cite{du2024cosyvoice2}, and EmoVoice \cite{yang2025emovoice}, enhance zero-shot performance using adaptive normalization or in-context tuning along with the LLMs. Nevertheless, most of them struggle to fully capture diverse emotional styles and stably remain timbre consistence from reference human speech \cite{qin2023openvoice}. These efforts mark significant steps toward emotion-aware expressive TTS but highlight the ongoing need for more generalizable human-level TTS synthesis frameworks \cite{tan2024naturalppeech}.

\subsection{Expressive Talking Head Generation}
Talking head generation aims at synthesizing realistic facial videos of a target identity synchronized with the driven audio~\cite{guo2021ad,shen2023difftalk}. 
Remarkable progress has been made in this task, benefiting from the rise of powerful generative techniques~\cite{mildenhall2021nerf,rombach2022high,kerbl20233d}. 
While most methods primarily focused on audio-lip synchronization~\cite{shen2022learning,guo2024rad,li2024talkinggaussian}, recent efforts have attempted to incorporate expressiveness into facial synthesis~\cite{danvevcek2022emoca,zhang2023sadtalker,tan2025dicetalk,wang2020mead,wu2021imitating}.  
MEAD~\cite{wang2020mead} introduces a large-scale emotional audio-visual benchmark and establishes a baseline for expressive talking head generation. LSF~\cite{wu2021imitating} and EMOCA~\cite{danvevcek2022emoca} rely on 3D Morphable Models for emotional facial control.
SadTalker~\cite{zhang2023sadtalker} disentangles structure and motion to enable expressive face generation. DICE-Talk~\cite{tan2025dicetalk} introduces a dynamic audio-expression co-modulation framework to bridge emotional semantics in speech with corresponding facial responses. 
Although these approaches emphasize emotion in talking head synthesis, they can hardly align the actual emotional tone of both textual, speech, and facial content, leading to perceptually unnatural or even contradictory results.

Building on previous practices, our empathetic response system infers emotions from the speech and dialogue context, explicitly enabling the synthesis of expressive talking speeches and videos that are emotionally aligned with the underlying semantics.

\section{E3RG System}
\label{sec:system}
\begin{figure*}[htbp]
	\centering 
	\includegraphics[scale=0.23] {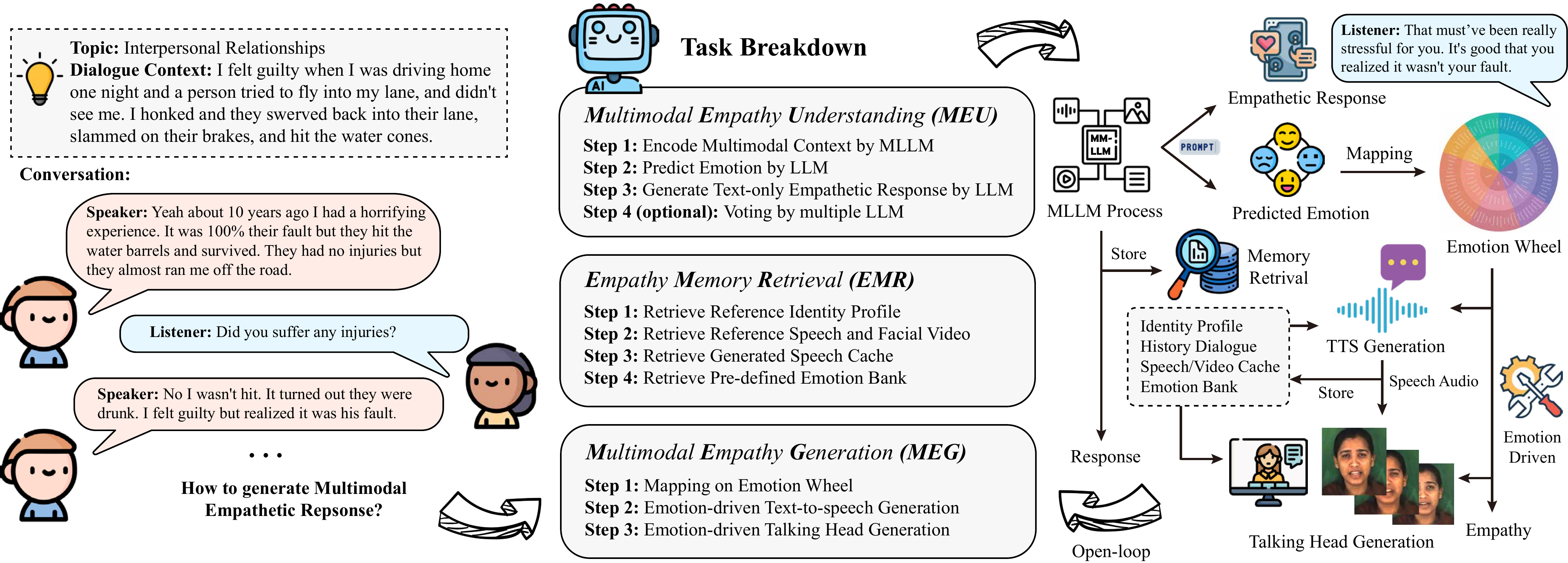}
	\caption{Overview of the proposed E3RG conversational system for multimodal empathetic response, consisting of empathy understanding, memory retrieval, and empathy generation sub-tasks.}
	\label{figure0}
    \vspace{-0.3cm}
\end{figure*}

This section presents the proposed E3RG system as shown in Figure \ref{figure0}. First, we define and breakdown Multimodal Empathetic Response Generation (MERG) task into several sub-tasks to reduce the difficulty and build a unified framework with flexible modules in a training-free manner (Sec. \ref{subsec:task}). Then, we leverage multimodal large language model to encode the multimodal context and conduct emotion prediction and empathetic text-based response (Sec. \ref{subsec:meu}). Next, we introduce a memory store and retrieval module to maintain the consistency of dialogue context, identity profile, or generated cache among different models (Sec. \ref{subsec:emr}). Lastly, we conduct text-to-speech and talking head generation guided by emotion and output the human-centric video as the final response (Sec. \ref{subsec:meg}).

\subsection{Task Definition and Breakdown}
\label{subsec:task}
MERG aims at constructing conversational systems between two avatars simulated as speaker and listener (human user and computer) to understand and produce the human-centric talking videos with rich empathy in multi-turn dialogue \cite{zhang2025avamerg}. Considering dialogue $\hat{D}=\{Q_i;D_{<i}\}$ where $D_{<i}=\{Q_{i-1},R_{i-1};Q_{i-2}, R_{i-2};...;Q_0,R_0\}$, the goal of MERG task is to output the corresponding response $R_i$ with the given user query $Q_i$ and the order historical dialogue $D_{<i}$. Each query or response is multimodal utterance $Q_i/R_i=\{L_i,A_i,V_i\}$ including triplet modalities as linguistic, speech, and facial videos, and $i\in[0,N]$ denotes the turn of dialogue where $N$ denotes the total number of dialogue turns for each sample. The MERG task not only focuses on generating natural multimodal content, but demands style consistency and emotional coherence across modalities, ensuring that empathetic cues are closely aligned among the generated response, user’s query and dialogue history.

Considering the complexity of the task, we first decompose the MERG task into three sub-tasks to more flexibly manage the conversation understanding and generation process, and enhance emotional and stylistic alignment across different modalities. The first sub-task is summarized as Multimodal Empathy Understanding (MEU), which leverages multimodal large language model to process the multimodal input, predict users' emotions, and generate a text-only empathetic response. The second sub-task is named as Empathy Memory Retrieval (EMR), where build memory bank to stores the identify profile, historical speech, and facial video, and the intermediate generated speech cache. The third sub-task is Multimodal Response Generation (MRG), which maps the predicted emotion to the pre-defined emotion bank and utilizes it to explicitly guide expressive text-to-speech or talking head generation. Conducting these three sub-tasks in a sequential interleaved execution manner as shown in Figure \ref{figure0}, the proposed conversational system can provide human-centric video with abundant emotion, aligning with the empathy demand of MERG task.

\subsection{Multimodal Empathy Understanding}
\label{subsec:meu}


\subsubsection{Multimodal Context Encoding with MLLM} 
Most of MLLMs jointly combine multimodal encoders and LLM to encode the multimodal content, where the multimodal encoders are utilized to process the audio or vision modalities, and LLM targets at understanding the textual modality. Considering triplet $\{L,A,V\}$ as text, audio, and vision modalities respectively, current multimodal encoding process of MLLM leverages a tokenizer to tokenize the text and concatenate the textual tokens with pre-processed acoustic or visual features \cite{fu2024mme}, formulated as:
    \begin{equation}
    \begin{aligned}
         F_L&=LLMTokenizer(L)\\
         F_a/F_v&=ModalityEncoder(A/V)
    \end{aligned}
    \end{equation}
with $Concat([F_L;F_a;F_v])$ serve as the input tokens of LLM. In this multimodal understanding task, the practical encoding process of MLLM can be divided into two types: 

\begin{itemize}
    \item[(1)] Connect separate diverse modality-specific models and LLM model, note that this type demands another finetuning process to align the semantic space among diverse modalities (such as using the audio-visual output by ImageBind \cite{girdhar2023imagebind} and input them with text together into any LLM);
    \item[(2)] Leverage Omni-Modal LLM to deal with multimodal context in a unified model (such as Qwen2.5-Omni \cite{jin2025qwen25omni}). Benefiting from the powerful cross-modal alignment built in the pre-training stage, this method can be used in a tuning-free way. 
\end{itemize}

Note that we utilize zero-shot or few-shot experiment settings in a training-free manner for LLMs in this paper, and leave the tuning process for future exploration.

\subsubsection{Emotion Prediction with LLM} 
After encoding multimodal content into $\{F_L;F_a;F_v\}$, we conduct a single-choice QA task on LLM to predict emotion contained in previous dialogues and the user's query. In practice, we ask the LLM to choose the most precise emotion in a pre-defined emotion set $EmoSet$, which can be flexibly modified as needed. The prompt template is presented as follows: 




% ...


\begin{tcolorbox}[breakable, colback=gray!10, colframe=black, title=Prompt template to predict emotion for MLLM:]
\vspace{-1.5mm}
\footnotesize{
Please act as an expert in the field of emotions. Please choose one most likely emotion from the given candidates for the speaker in the given dialogue: \begin{verbatim}
EmoSet = {neutral, happy, surprised, angry, fear, 
sad, disgusted, contempt...}
\end{verbatim}
Respond with only one word for the chosen emotion. Do not include any other text. 
\textbf{The dialogue is:} 
\begin{verbatim}
Speaker: "string" <Aud> <Vid> \n
Listener: "string" <Aud> <Vid> \n
...
Speaker: "string" <Aud> <Vid> \n
\end{verbatim}
The emotion class of the Speaker:
}
\vspace{-1.5mm}
\end{tcolorbox}

Here <$Aud$> and <$Vid$> are the special placeholders which will be replaced with corresponding audio and visual features, and latter be concatenated with the token embeddings of $"string"$ and prompt text before being fed into LLM.

\subsubsection{Text-only Empathetic Response Generation with LLM} Given the following prompt template, we utilize the same LLM as the one predicting emotion to generate text-only empathetic response. Note that the prompt can be replaced into CoT-type prompt \cite{zhang2025avamerg}, and we can further tune the LLM on specific datasets to enhance the empathetic accuracy of textual responses.

\begin{tcolorbox}[breakable, colback=gray!10, colframe=black, title=Prompt template to predict response for MLLM:]
\vspace{-1.5mm}
\footnotesize{
Please act as an empathetic responser. Please output the listener's next response to the speaker in the given dialogue. Note that the response should show the concern of listener and attempting to address the speaker’s emotional state. \\
Output the response directly. Do not include any other words. 
\textbf{The dialogue is:} 
\begin{verbatim}
Speaker: "string" <Aud> <Vid> \n
Listener: "string" <Aud> <Vid> \n
...
Speaker: "string" <Aud> <Vid> \n
\end{verbatim}
The response of the Listener:
}
\vspace{-1.5mm}
\end{tcolorbox}

\subsubsection{Voting with multiple LLMs} 
Since LLMs have been empirically validated with inherent affective bias and unsound determine capability when conducting tasks about emotion intelligence \cite{mao2023bias,zhao2023chatgpt}, we remain an optional step to leverage multiple LLMs to predict emotion and response at the same time. By conducting voting strategy \cite{yang2024llmvoting,chen2024more}, we can further improve the accuracy of emotion prediction and obtain a more empathetic response. Considering $\hat{E}_k$ as the predicted emotion and $\hat{R}_k$ as the generated response of $k$-th LLM, the proposed system contains two kinds of voting strategies as follows:

\begin{itemize}
    \item[(1)] Leverage majority voting to choose the most selected emotion class from the results of all LLMs, and utilize the corresponding empathetic response output by the same LLM.
    \item[(2)] Conduct weighted voting on the output from all LLM where the weight can be obtained by the emotional intelligence performance of each LLM, and then choose the emotion class with the highest score. The response from the same LLM is also selected as the final response. 
\end{itemize}

Note that we utilize the majority voting strategy in the experiments of this paper for simplicity, and leave the weighted voting strategy for future work.

\subsection{Empathy Memory Retrieval}
\label{subsec:emr}


\subsubsection{Reference Identity Profile Retrieval}
Each speaker and listener profile, including attributes such as age, gender, and vocal timbre, directly influences the tone and style of their responses. For instance, a child’s speech might be characterized by a higher pitch and immature tune, whereas an adult’s response could contain a deeper timbre and a measured pace. Furthermore, leveraging additional dialogues from the same individual helps the system maintain a consistent speaking style and more accurately capture their emotional nuances when generating responses. The identity profile is represented in JSON format as follows:

\begin{tcolorbox}[breakable, colback=gray!10, colframe=black, title=Identify Profile:]
\vspace{-1.5mm}
\footnotesize{
\textbf{JSON Format Example:}
\begin{verbatim}
{ 
  "speaker_profile"/"listener_profile": { 
      "ID": "int", "age": "string",
      "gender": "string", "timbre": "string", 
      "reference_utterance": "path_string", 
      "reference_speech": "path_string", 
      "reference_facial": "path_string" 
  }
}
\end{verbatim}
}
\vspace{-1.5mm}
\end{tcolorbox}

\subsubsection{Reference Speech and Facial Video Retrieval}
To ensure identity consistency in generated responses, we further retrieve past speech and facial video frames of the relevant speaker or listener. These retrieved samples then serve as the reference audio and visual anchors during the text-to-speech and talking‑head generation stages. By grounding generation in authentic and identity-specific cues, we guarantee that each multimodal response aligns seamlessly with that individual’s prior dialogues, which preserves both vocal characteristics and facial appearance for a coherent conversational persona-based avatar.

\subsubsection{Generated Speech Cache Retrieval}
Since the video response generation process involves two stages, including speech generation followed by video synthesis, the output of the first stage (i.e., the generated speech) should be temporarily stored. To support this, the generated speech is cached and later retrieved as input during the talking head generation stage, ensuring efficient and seamless multimodal video synthesis.

\subsubsection{Pre-defined Emotion Bank Retrieval}
As previously mentioned, emotion plays a crucial role in generating expressive and empathetic responses. To this end, we introduce the pre-defined emotion bank used for both speech and talking-head synthesis \cite{qin2023openvoice,tan2025dicetalk}. After predicting the emotional state of dialogues with LLM, the system selects the corresponding emotion embedding or token from the emotion bank. This retrieved emotion prior is then incorporated into the generation pipeline, enabling the model to produce emotionally aligned and empathetic multimodal responses.

\subsection{Multimodal Empathy Generation}
\label{subsec:meg}


\subsubsection{Mapping on Emotion Wheel}
Considering the fine-grained emotion classes predicted by the LLM, we incorporate the Emotion Wheel \cite{plutchik1980emotion} to map these detailed emotions into coarser or semantically similar categories that align with the pre-defined emotion classes in the speech or video generation emotion banks. This mapping stage not only bridges the gap between nuanced emotional understanding and model-executable conditioning, but also enhances the versatility and transferability of the proposed system. By aligning predicted emotions with standardized categories in the emotion wheel, the system can seamlessly adapt to future upgrades or post-training refinements of expressive generative models.

\subsubsection{Emotion-driven Text-to-Speech Translation}
After generating the textual response using the LLM, we employ an expressive TTS model to synthesize speech that reflects both the response content and the predicted emotion. To ensure identity consistency between the transcribed response and previous dialogues, the voice characteristics of the talker are preserved in our system. For this purpose, we adopt  OpenVoice \cite{qin2023openvoice}, which strikes a balance between computational efficiency and naturalness in zero-shot voice cloning. In practice, a base speaker model is used to control speaking styles and language, while a converter model transfers the timbre of the reference audio to the translated speech. Notably, emotional cues are embedded within the speaking style, encompassing features such as accent, rhythm, pauses, and intonation. The available speaking emotion style is presented as follows:

\begin{tcolorbox}[breakable, colback=gray!10, colframe=black, title=Emotion Bank for TTS Translation:]
\vspace{-1.5mm}
\footnotesize{
\begin{verbatim}
  Speaking Style = {friendly, cheerful, excited, sad, 
  angry, terrified, shouting, whispering}
\end{verbatim}
}
\vspace{-1.5mm}
\end{tcolorbox}

\subsubsection{Emotion-driven Talking Head Generation}
Finally, we employ the state‑of‑the‑art audio‑driven talking‑head generator, DICE‑Talk \cite{tan2025dicetalk}, to produce emotionally nuanced video portraits while rigorously preserving speaker identity. The generator represents each emotion as an identity‑agnostic Gaussian distribution, effectively preventing identity leakage and leveraging speech prosody as a natural emotional cue. Specifically, a cross‑modal emotion embedder disentangles emotion semantics from individual identities and captures inter‑emotion relationships between facial movements and vocal expressions. Guided by the emotion prior, the model then combines historical facial images with the translated speech in a synergistic manner to animate lifelike talking heads that faithfully reflect both the talker’s unique appearance and their intended emotional state. As a result, our system delivers realistic video responses that maintain multimodal identity consistency and rich empathy.

\begin{tcolorbox}[breakable, colback=gray!10, colframe=black, title=Emotion Bank for Talking Head Generation:]
\vspace{-1.5mm}
\footnotesize{
\begin{verbatim}
  Facial Emotion = {angry, contempt, disgusted, fear, 
  happy, sad, surprised, neutral}
\end{verbatim}
}
\vspace{-1.5mm}
\end{tcolorbox}

\section{Experiment}

\begin{figure*}[htbp]
    \centering 
    \subfloat[Neutral] {
    \includegraphics[width=0.5\columnwidth]{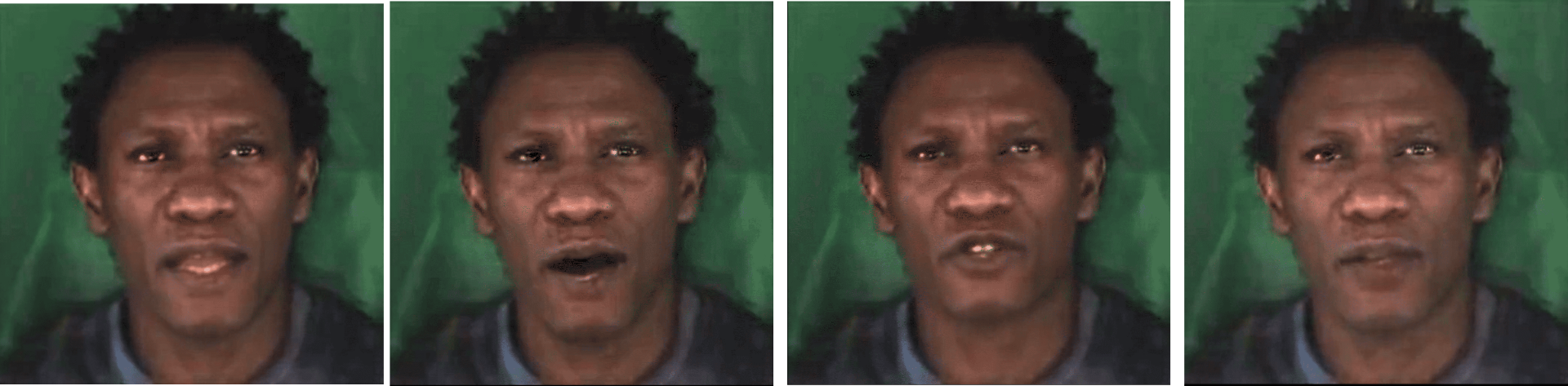}
    } 
    \subfloat[Happy] {
    \includegraphics[width=0.5\columnwidth]{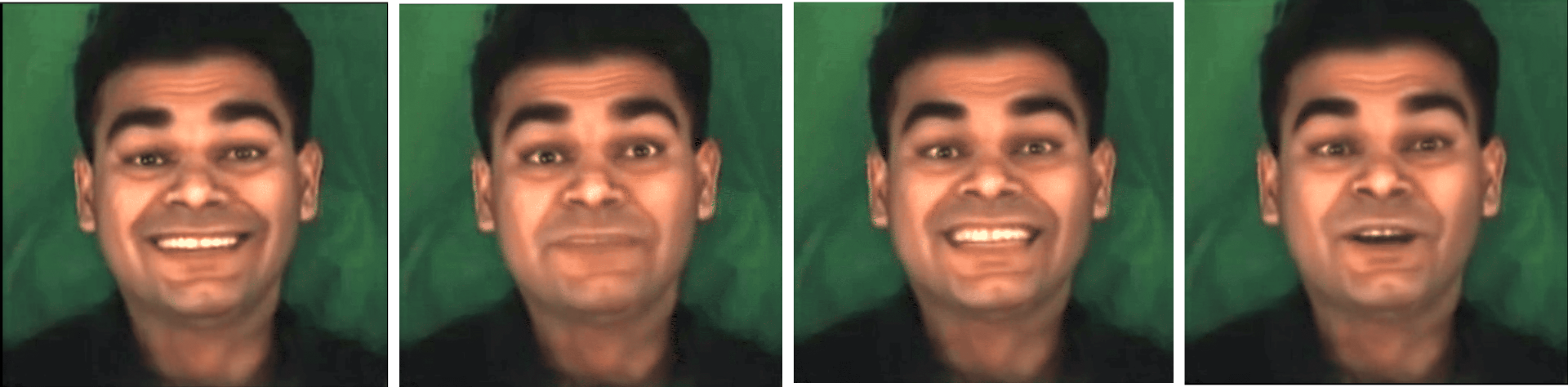}
    } 
    \subfloat[Fear] {
    \includegraphics[width=0.5\columnwidth]{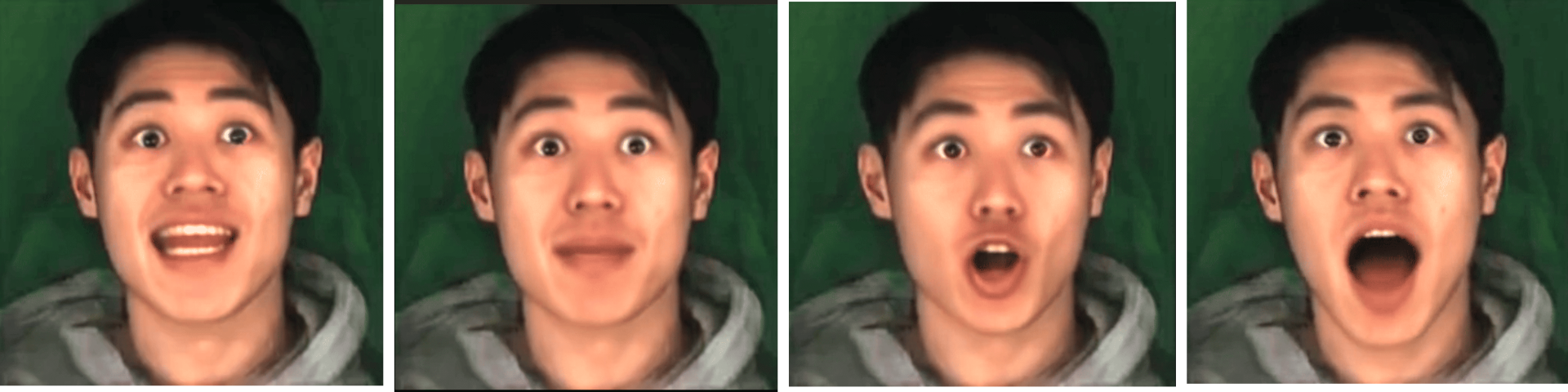}
    } 
    \subfloat[Angry] {
    \includegraphics[width=0.5\columnwidth]{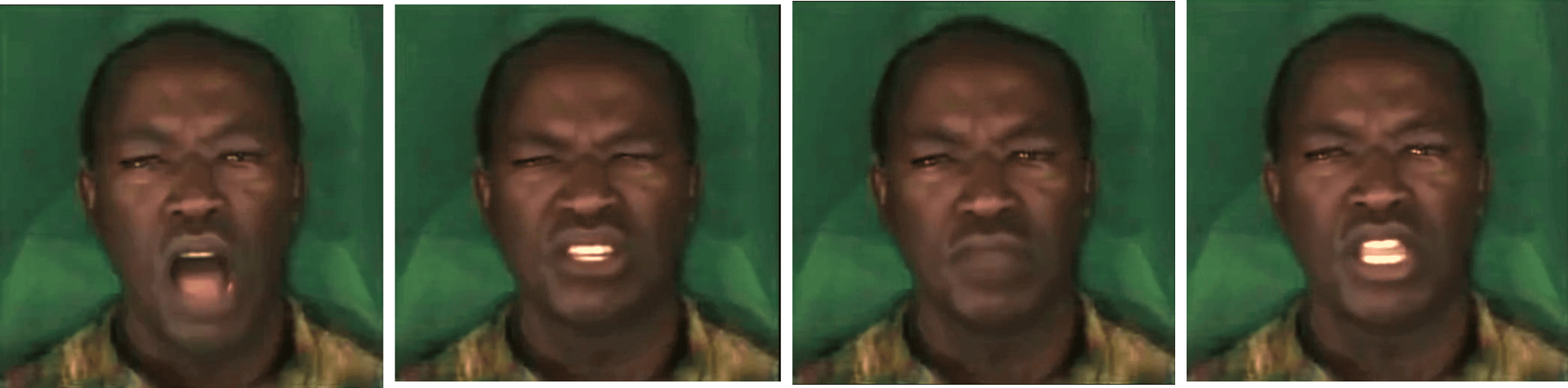}
    } 
    \\
    \vspace{-0.2cm}
    \subfloat[Disgusted] { 
    \includegraphics[width=0.5\columnwidth]{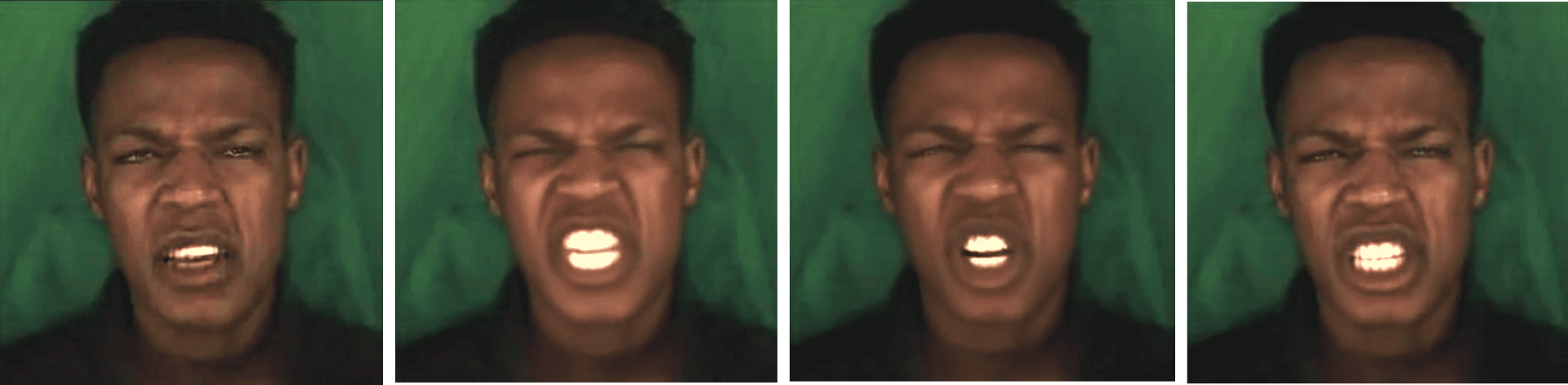}     
    }
    \subfloat[Sad] { 
    \includegraphics[width=0.5\columnwidth]{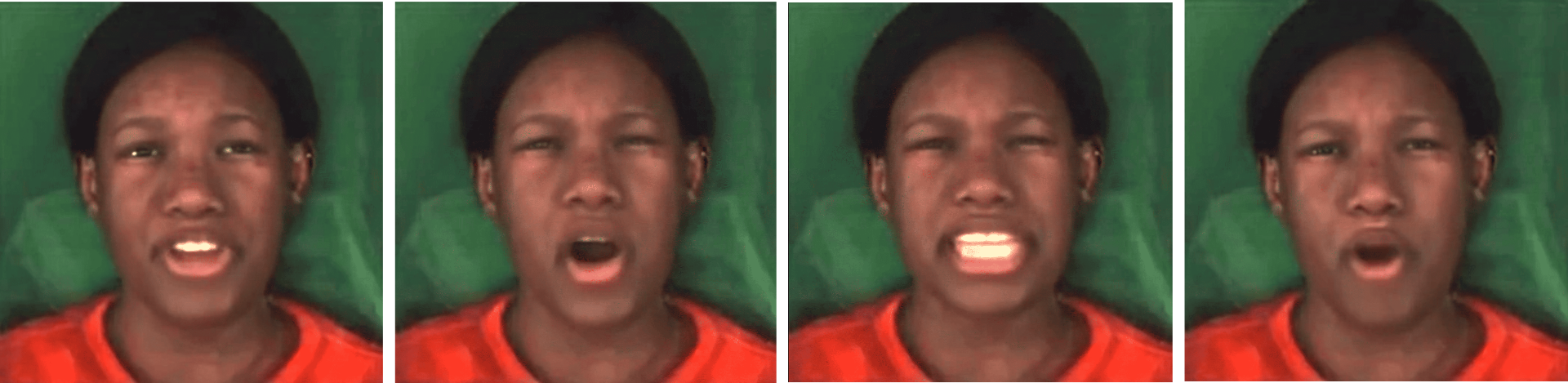}
    }
    \subfloat[Surprised] { 
    \includegraphics[width=0.5\columnwidth]{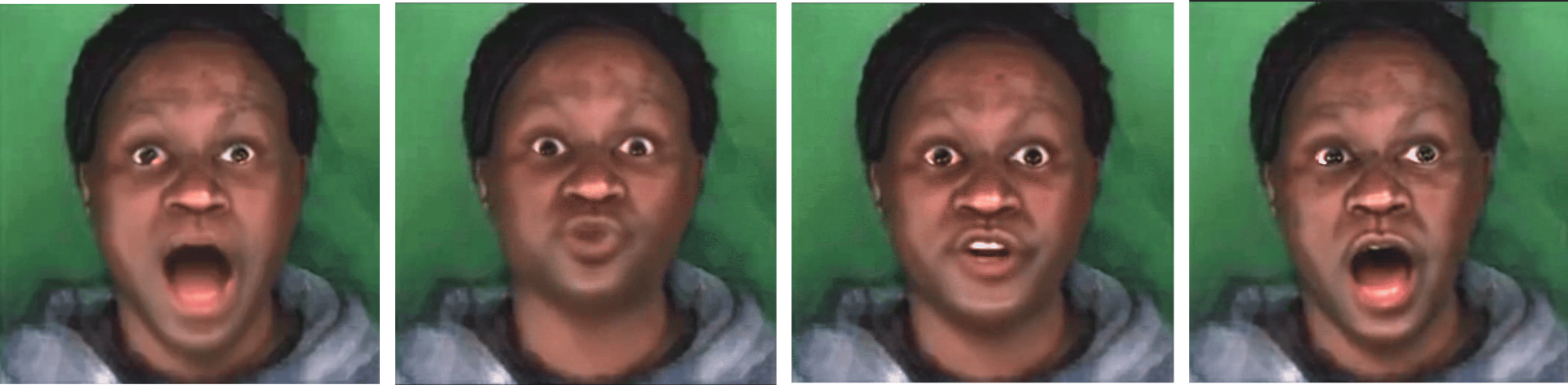}
    }
    \subfloat[Contempt] { 
    \includegraphics[width=0.5\columnwidth]{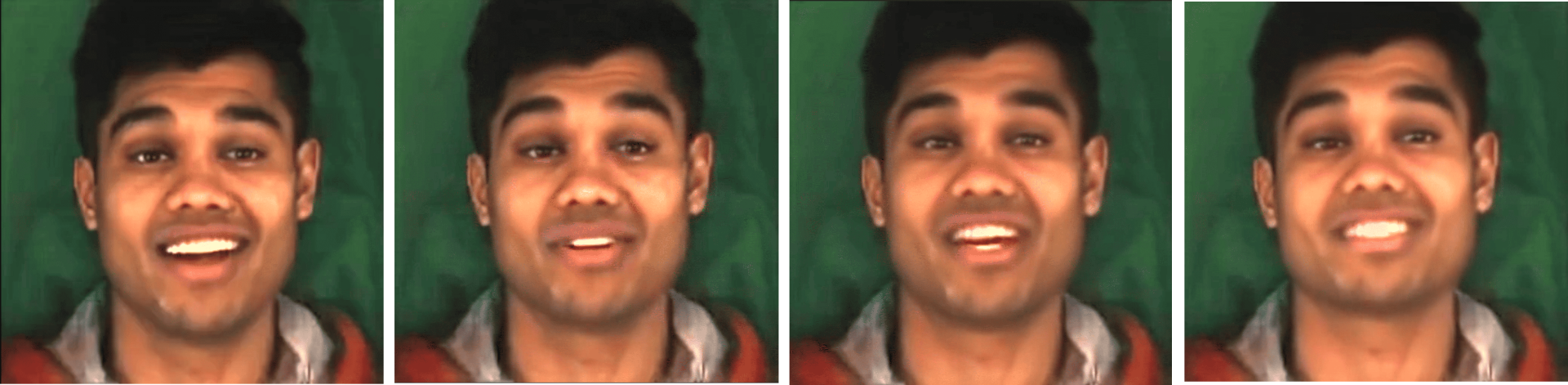}
    }
    \vspace{-0.3cm}
    \caption{Visualization of zero-shot qualitative results with the guidance of diverse emotion classes.}
    \label{vis_emotion}
\vspace{-0.3cm}
\end{figure*}
\subsection{Dataset}
\textbf{AvaMERG}~\cite{zhang2025avamerg} is a large‑scale, multimodal empathetic dataset comprising 33,048 dialogues and 152,021 utterances, built upon the \textsc{EmpatheticDialogues} corpus~\cite{rashkin2019empatheticdialogues}. Each dialogue includes aligned text, speech, and avatar video, and is categorized into 10 primary topics and hundreds of fine-grained subtopics reflecting common real‑world scenarios. The dataset covers 7 emotions (happy, fear, angry, disgusted, sad, surprised, and contempt) and provides rich annotations to support the development of MERG systems.

\subsection{Evaluation Metric}
\textbf{Dist-n} \cite{li2016distn} is computed to measure the diversity for the LLM-generated textual responses and \textbf{HIT} Rate ($\%$) \cite{lian2025affectgpt} is adopted to evaluate the emotion prediction accuracy, implicitly indicating model's empathetic ability. Besides, human evaluation \cite{zhang2025avamerg} is conducted on the generated video responses in three aspects: \textbf{Emotional Expressiveness} evaluating how the response conveys emotions through facial expressions, vocal tone, and the corresponding empathetic text; \textbf{Multimodal Consistency} validating the consistency of verbal, facial, and vocal expressions; and \textbf{Naturalness} capturing human-like degree the response appears.


\subsection{Quantitative Results}
We present the proposed E3RG system equipped with various LLMs and MLLMs, as detailed in Table \ref{exp_llm}. In zero-shot setting, MiniCPM4 and Ola-Omni achieves the highest performance on both emotion prediction and response diversity. Under few-shot setting \cite{qian2023harnessing} randomly sampling $n$ instances as examples in prompt, further improvements on emotion understanding are observed. The experiment results indicate the broad applicability of the proposed approach across different models without additional training. Besides, the comparison between text-only and omni-modal LLMs highlights the advantages of incorporating multimodal context. Moreover, the human evaluation results in Table \ref{table_challenge} present the effectiveness of the E3RG system, surpassing other teams on average score. 

\begin{table}[htbp]
\caption{Comparison of the zero-shot (no extra mention) and few-shot ($n$-shot) performance on the training set of AvaMERG dataset for the state-of-the-art LLM and MLLM.}\label{exp_llm}
\centering
\scalebox{0.8}{
\setlength\tabcolsep{5pt}
\begin{tabular}{c|c|ccc}
\hline
  \multicolumn{2}{c|}{LLM/MLLM Model} & HIT & Dist-1 & Dist-2 \\ 
\hline
  \multirow{9}{*}{\makecell{Text-only\\LLM}} & Vicuna-1.5-7B~\cite{chiang2023vicuna} & 46.0 & 0.825 & 0.960 \\
  & Llama-3-8B~\cite{grattafiori2024llama3} & 59.4 & 0.849 & 0.985 \\
  & InternLM3-8B~\cite{cai2024internlm2}  & 65.3 & 0.943 & 0.997 \\
  & Qwen2.5-7B~\cite{qwen2024qwen25} & 69.3 & 0.967 & 0.999 \\
  & Qwen2.5-7B (1-shot) & 70.7 & 0.977 & 0.999 \\
  & Qwen2.5-7B (3-shot) & 73.2 & 0.978 & 0.999 \\
  & MiniCPM4-8B~\cite{team2025minicpm4} & 73.9 & 0.983 & 0.999 \\
  & MiniCPM4-8B (1-shot) & \textbf{74.7} & 0.984 & 0.999 \\
  & MiniCPM4-8B (3-shot) & 74.2 & \textbf{0.985} & 0.999\textbf{} \\
\hline
  \multirow{5}{*}{\makecell{Omni-Modal\\LLM}} 
  & MiniCPM-o 2.6 8B~\cite{yao2024minicpmv} & 65.8 & 0.952 & 0.996 \\
  & Qwen2.5-Omni-7B~\cite{jin2025qwen25omni} & 72.3 & 0.986 & 0.997 \\
  & Ola-Omni-7B~\cite{liu2025ola} & 75.6 & 0.986 & 0.998 \\
  & Ola-Omni-7B (1-shot) & 76.1 & 0.989 & 0.999 \\
  & Ola-Omni-7B (3-shot) & \textbf{76.3} & \textbf{0.990} & 0.999 \\
\hline
\end{tabular}
}
\vspace{-0.3cm}
\end{table}

\subsection{Qualitative Results}
As shown in Figure \ref{vis_emotion}, we showcase the generated empathetic videos exhibiting a range of emotions through facial expressions guided by various emotions. The visualization results demonstrate not only strong identity consistency within each multimodal response, but also the system's ability to capture emotionally rich content with natural and lifelike facial appearance and movement.





\begin{table}[htbp]
\caption{Comparison with human evaluation performance for responses generated by different competition teams on the testing set of AvaMERG dataset. }\label{table_challenge}
\centering
\scalebox{0.8}{
\setlength\tabcolsep{5pt}
\begin{tabular}{c|ccc|c}

\hline
\multicolumn{1}{c|}{Team} & \begin{tabular}[c]{@{}c@{}}Emotional\\Expressiveness\end{tabular} &
\begin{tabular}[c]{@{}c@{}}Multimodal\\Consistency\end{tabular} & Naturalness & Average \\

\hline


  It's MyGO & 3.5 & 3.5 & 3.2 & 3.40 \\
  AI4AI & 3.6 & 3.8 & \textbf{4.1} & 3.83 \\
  \textbf{Ours} & \textbf{4.3} & \textbf{4.0} & 3.8 & \textbf{4.03} \\
\hline
\end{tabular}
}
\vspace{-0.3cm}
\end{table}

\section{Conclusion}
In this paper, we introduced E3RG system by dividing the MERG task into three sub-tasks, adopting emotion guidance and identity consistency in MLLM understanding and expressive speech and video generative models. E3RG produces natural and emotionally rich responses across text, speech, and video in a training-free manner. The proposed system  reaches the highest performance in both automatic and human evaluations, achieving Top-1 position in the Avatar-based Multimodal Empathy Challenge on ACM MM’25.


\begin{acks}
This work was supported by the National Natural Science Foundation of China (62076262, 61673402, 61273270, 60802069) and by the International Program Fund for Young Talent Scientific Research People, Sun Yat-sen University.
\end{acks}

\bibliographystyle{ACM-Reference-Format}
\balance
\bibliography{empathyresponse}




\end{document}